\documentclass[12pt,journal,onecolumn,draftclsnofoot]{IEEEtran}
\usepackage{dsfont}
\usepackage{amsmath,amsfonts,bm,amssymb}
\usepackage{mathtools}
\usepackage{xcolor}
\usepackage{graphicx}
\usepackage{empheq}

\usepackage{amsmath,amsfonts,bm}

\def\secref#1{section~\ref{#1}}
\def\Secref#1{Section~\ref{#1}}

\def\eqref#1{equation~\ref{#1}}

\def\Algref#1{Algorithm~\ref{#1}}

\def\Theoref#1{Theorem~\ref{#1}}

\def\1{\bm{1}}

\def\valpha{{\bm{\alpha}}}

\def\evalpha{{\alpha}}

\def\evmu{{\mu}}

\def\mS{{\bm{S}}}

\def\mW{{\bm{W}}}
\def\mX{{\bm{X}}}

\DeclareMathAlphabet{\mathsfit}{\encodingdefault}{\sfdefault}{m}{sl}
\SetMathAlphabet{\mathsfit}{bold}{\encodingdefault}{\sfdefault}{bx}{n}

\def\gA{{\mathcal{A}}}

\def\gE{{\mathcal{E}}}

\def\gH{{\mathcal{H}}}

\def\sM{{\mathbb{M}}}

\def\sP{{\mathbb{P}}}

\def\sR{{\mathbb{R}}}
\def\sS{{\mathbb{S}}}

\def\sZ{{\mathbb{Z}}}

\def\emS{{S}}

\usepackage{algorithm}
\usepackage[noend]{algpseudocode}
\usepackage{booktabs}
\usepackage{threeparttable}
\usepackage{soul}
\usepackage{thmtools,thm-restate}
\usepackage{ulem}

\newtheorem{claim}{Claim}
\usepackage{caption}
\usepackage[colorlinks=true, citecolor=blue]{hyperref}

\newcommand{\violet}{\textcolor{black}}
\newcommand{\lin}[1]{\textcolor{red}{[Lin: #1]}}

\newcommand{\algoname}{\text{BatchSP2 }}

\newcommand*\widefbox[1]{\fbox{\hspace{2em}#1\hspace{2em}}}

\newif\ifarxivFormat
\arxivFormattrue

\begin{document}

\title{Multi-Agent Bandit Learning through \\ Heterogeneous Action Erasure Channels} 

\author{\IEEEauthorblockN{Osama A. Hanna$^\dagger$*, Merve Karakas$^\dagger$*, Lin F. Yang$^\dagger$, and Christina Fragouli$^\dagger$ \\ 
$^\dagger$University of California, Los Angeles\\
Email:\{ohanna, mervekarakas, linyang, christina.fragouli\}@ucla.edu}
\ifarxivFormat\thanks{
* indicates equal contribution. This work is partially supported by NSF grants \#2007714 and \#2221871, by Army Research Laboratory grant under Cooperative Agreement W911NF-17-2-0196, and by Amazon Faculty Award.
}\fi
}
\maketitle
\allowdisplaybreaks
\begin{abstract}
    Multi-Armed Bandit (MAB) systems are witnessing an upswing in applications within multi-agent distributed environments, leading to the advancement of collaborative MAB algorithms. In such settings, communication between agents executing actions and the primary learner making decisions can hinder the learning process. A prevalent challenge in distributed learning is action erasure, often induced by communication delays and/or channel noise. This results in agents possibly not receiving the intended action from the learner, subsequently leading to misguided feedback. In this paper, we introduce novel algorithms that enable learners to interact concurrently with distributed agents across heterogeneous action erasure channels with different action erasure probabilities. We illustrate that, in contrast to existing bandit algorithms, which experience linear regret, our algorithms assure sub-linear regret guarantees. Our proposed solutions are founded on a meticulously crafted repetition protocol and scheduling of learning across heterogeneous channels. To our knowledge, these are the first algorithms capable of effectively learning through heterogeneous action erasure channels. We substantiate the superior performance of our algorithm through numerical experiments, emphasizing their practical significance in addressing issues related to communication constraints and delays in multi-agent environments.
\end{abstract}

\section{INTRODUCTION}
\label{intro}
 Multi-armed bandits, a well-established and effective online learning model, are increasingly finding applications in multi-agent distributed environments. One notable use-case involves leveraging a central learner, with actions (arms) communicated to remote agents to collect rewards, as discussed in \cite{hanna2022solving, hanna2022learning, hanna2023compression}. However, a noteworthy gap in the existing literature pertains to scenarios where the communicated actions may be lost due to communication channel issues such as delays or noise interference. This challenge becomes even more pronounced when various agents possess communication channels with varying capabilities, and these channels do not provide feedback regarding the receipt of actions. Throughout this paper, we will refer to ``feedback" to denote receipt acknowledgments from the channel, distinguishing this from the rewards, which represent feedback on the learned actions.


This challenge has not been well explored in the literature, as most works assume that agents will acknowledge whether an action request has been received or not.  Yet the assumption of feedback availability can have high cost or  simply not be possible.
For instance, distributed recommendation systems may send content (action requests)  over wireless channels that are notoriously subject to delays due to varying channel conditions and lost packets \ifarxivFormat\cite{kurose2012computernetworks} \else\citep{kurose2012computernetworks}\fi; even wired networks are subject to significant delay variability due to factors such as network topology, queuing delay and prioritization within cloud databases \ifarxivFormat\cite{yeh2014vipnetwork, dehghan2019sharingnetwork,kurose2012computernetworks} \else\citep{yeh2014vipnetwork, dehghan2019sharingnetwork,kurose2012computernetworks}\fi. Meanwhile, even if the content is delivered and displayed to the agents (e.g., an app recommending to follow a route, visit a restaurant, etc.), we cannot be sure when exactly a human user sees it, if at all.  

Another motivating case is when the agents are devices with very limited communication capabilities. 
One such application is  fleets of medical micro-robots
(which today can be even of  nanometre-scale)  that  propel themselves through biological media, such as the veins and the gastrointestinal tract \ifarxivFormat\cite{microswimmers2023, zou2022gaitmicroswimmer, amir2020prediction} \else\citep{microswimmers2023, zou2022gaitmicroswimmer, amir2020prediction}\fi. Multi-agent MAB algorithms can facilitate personalizing the robot's actions to different patients, for instance, to release tailored amounts of substances or to attack specific particles. The rewards (capturing action outcomes) are usually observed through external medical equipment, such as ultrasound or other imaging; however, conveying what action to play to the robots, can be communication challenging. 
A third case is that of military operations, where a central commander may want to communicate actions to agents  (such as small robots), who do not wish to communicate back so as not to reveal their position in a hostile territory, yet their actions may have impact observable through  satellite imaging or sensors.


In this work, we dispense with the need for feedback. We ask, what performance can we achieve if the learner action requests are delivered according to a known probabilistic model, but we have no additional information on whether each specific request is delivered or not. In particular, we assume operation over $T$ rounds, where at each round a central learner sends commands (which action to play) to $M$ agents through  erasure channels with erasure probability $\epsilon_i$, $i=1\ldots M$, where these probabilities can be arbitrarily different across channels. This induces a \violet{Geometric} distribution on 
the reception time of each action request, different for each agent\footnote{We note that our schemes extend to more general such probability distributions; \violet{see \Secref{sec:delay_interpretation} for a detailed discussion.}}.
The agents send no feedback (thus the learner does not know which action request the agents are following); 
the agents play at each round the last action command they received.
The learner observes the reward for the (uknown) action  played through an error-free channel - which can lead to erroneous  action-reward associations. Indeed, energy and space limitations of micro-robots, security constraints in military applications, or simply the structure of the communication protocol can prevent transmitting feedback back to the learner; and agents are required to perform an action at each time, since even no action (for instance staying still) is also an action, see Table~\ref{table:system-model-example} for a small example. 


Our objective is to design multi-agent MAB schemes that minimize the impact of action erasures on the  regret, while also leveraging the multi-agent setting to expedite the learning process.
Our main contributions are as follows:
\begin{itemize}


    \item We propose \algoname, a Successive Arm Elimination based repetition algorithm with a crafted scheduling part for multi-agent MAB setup with erasures, and prove sub-linear regret guarantees on the proposed algorithm.
    \item We provide numerical comparisons with a number of baseline algorithms, show the superiority of our algorithm to the benchmarks and that simply applying existing MAB algorithms in a manner oblivious to action erasures can lead to linear regret.
\end{itemize}

\paragraph{Related Work.} \label{related_work}

Various MAB algorithms achieving optimal or near-optimal regret bounds under different assumptions have been proposed over the years \ifarxivFormat\cite{lattimore2020bandit}\else\citep{lattimore2020bandit}\fi. Previous studies designed optimal algorithms for the simple MAB setting through providing gap dependent regret of $\tilde{O}(\sum_{a:\Delta_a>0}\frac{1}{\Delta_a})$ and worst-case regret of $O(\sqrt{KT\log{T}})$ \ifarxivFormat\cite{Thompson1933ONTL, Auer2002b, Lai1987AdaptiveTA} \else\citep{Thompson1933ONTL, Auer2002b, Lai1987AdaptiveTA}\fi. However, these algorithms are not resilient to action erasures and they are not optimized for multi-agent settings. \violet{Building upon MAB algorithms, there has been a considerable amount of recent work on multi-agent MABs in various settings \ifarxivFormat\cite{7952664-mamab-playerdependentreward, dubey2020cooperative-mamab, 10.5555/3586589.3586801-mamab-limitedcomm, xu2023decentralized-mamab}\else\citep{7952664-mamab-playerdependentreward, dubey2020cooperative-mamab, 10.5555/3586589.3586801-mamab-limitedcomm, xu2023decentralized-mamab}\fi. However, these works predominantly consider connected agents with some form of communication between neighbors or improve the regret bounds through feedback mechanisms, primarily involving collision sensing for agents \ifarxivFormat\cite{pmlr-v108-wang20m-optmpmab, shi2021heterogeneous-mamab}\else\citep{pmlr-v108-wang20m-optmpmab, shi2021heterogeneous-mamab}\fi. Consequently, they fail to effectively handle action erasures, particularly when connections between agents and feedback mechanism are missing.}


Taking a step back, our work fits within the framework of heterogeneous distributed agents supporting central  learning. The heterogeneity in our setup comes from the diversity in the communication channels (different erasure probabilities). Although as we discuss next
several works have considered  heterogeneous setups, communication channel diversity and how it can affect MAB learning is we believe a natural setup that has not been widely explored.

Multi-armed bandits with delayed feedback has been studied in recent years under different settings due to its practical applications. For the stochastic setting, \cite{pmlr-v28-joulani13} shows that bounded unknown i.i.d. delays cause an additive increase in the regret, i.e., $O(\sqrt{ KT \log{T}} + K \mathbb{E} \left[ D \right])$ where the first term is the regret of stochastic MAB problem with no delays at round $T$, $K$ is the number of actions, and $\mathbb{E}\left[D\right]$ is the expected delay. Following the work in \ifarxivFormat\cite{pmlr-v28-joulani13}, \cite{Mandel_Liu_Brunskill_Popović_2015} \else\citet{pmlr-v28-joulani13}, \citet{Mandel_Liu_Brunskill_Popović_2015} \fi  proposes a queue-based MAB algorithm to handle delays. Later, \ifarxivFormat\cite{pike2018bandits} \else\citet{pike2018bandits} \fi  achieves the same additive increase in regret as in \ifarxivFormat\cite{pmlr-v28-joulani13} \else\citet{pmlr-v28-joulani13} \fi under delayed aggregated anonymous feedback. \ifarxivFormat\cite{vernade2017stochastic} \else\citet{vernade2017stochastic} \fi studies Bernoulli bandits with known delay distribution where some feedback could also be censored, i.e., do not reach the learner. Relevant to these works, \ifarxivFormat\cite{pmlr-v84-grover18b} \else\citet{pmlr-v84-grover18b} \fi proposes an algorithm for the best arm identification problem in stochastic MABs with partial and delayed feedback where the aim is to minimize the number of samples for identifying the best action. They extend their methods to the parallel MAB setting, i.e., multiple actions are pulled at each time; however, they provide no lower bounds on the sample complexity of their problem setting. While these works incorporate delays into the stochastic MAB model, delays are associated with action pulls whereas, \violet{in our setup, delays are associated with agents and are independent on the pulled action for the same agent.}

A recent work considers the single agent action erasure channel \ifarxivFormat\cite{hanna2023multi} \else\citep{hanna2023multi} \fi and they provide a generic repetition scheme that works on top of any MAB algorithm and gets regret at most $O(1/\sqrt{1-\epsilon})$ away, and a specific algorithm that gets $O(\sqrt{KT} + K / \sqrt{1-\epsilon})$ regret that is near optimal; our model accepts their work as a special case; however, extending their methods to our case is highly non-trivial, 
as the main challenge being the variability between erasure probabilities of each channel, that induces a need for careful scheduling across agents.

\paragraph{Paper Organization} In \Secref{problem_formulation}, we introduce the notation and system model; we explain the proposed algorithm in \Secref{algorithm}; analyze it in \Secref{sec:analysis} and provide upper bounds; evaluate and compare with possible baselines in \Secref{experiments}; and conclude in \Secref{conclusion}.

\section{PROBLEM FORMULATION}
\label{problem_formulation}



\subsection{Multi-armed Bandits} We consider a stochastic multi-armed bandit problem in which a learner plays an action $a_t \in \gA$ at each round $t$ from the set of possible actions $\mathcal{A}$ and receives a reward $r_t$ associated with the played action. This interaction is repeated over a horizon $T$, i.e., $t \in \left\{1, 2, ..., T \right\}$ and the learner aims to maximize the cumulative reward at the end of $T$ rounds. The set of possible actions $\gA$ are the same throughout the horizon and have $K$ elements, i.e., $|\gA| = K$. The decision of the learner on which action to play may depend on the history $\gH_t = \left\{ a_1, r_1, a_2, r_2, ..., a_{t-1}, r_{t-1}\right\}$. Additionally, in a stochastic setting, the reward for each action $a$ is generated from an unknown reward distribution with an unknown mean $\mu_a$. In our analysis, we assume that the rewards are in the interval $\left[0, 1\right]$; however, our results directly extend to sub-Gaussian distributions. The objective of the learner is minimize the regret  over a time horizon $T$  defined as 
\[
    R_T = T \max_{a \in \gA} \mu_a - \mathbb{E}\left[ \sum_{t=1}^{T} r_t \right]
\]

\setlength{\tabcolsep}{6pt} 
\begin{table*}[tbhp!]
\centering
\begin{threeparttable}
\caption{Example of a MA-MAB Learning Over Action Erasure Channels. 
At each time $t$ the learner sends action requests to each agent; an agent that does not receive the request, simply continues to play the last received action. } \label{table:system-model-example}

\begin{tabular}{lcccccc}
\textbf{}  & \textbf{t=1} & \textbf{t=2} & \textbf{t=3} & \textbf{t=4} & \textbf{t=5} & \textbf{...} \\
\toprule
\textbf{Learner}         & $\{a_1^{(m)}\}_{m=1}^{M}$ & $\{a_2^{(m)}\}_{m=1}^{M}$ & $\{a_3^{(m)}\}_{m=1}^{M}$ & $\{a_4^{(m)}\}_{m=1}^{M}$ & $\{a_5^{(m)}\}_{m=1}^{M}$ & $\cdots$\\

\midrule
Erasure ($\epsilon_1=0.1$) & & & &X & &$\cdots$ \\
\textbf{Agent 1} ($\Tilde{a}_{t}^{(1)}$) &$a_1^{(1)}$  & $a_2^{(1)}$ & $a_3^{(1)}$ & $a_3^{(1)}$ & $a_5^{(1)}$ &$\cdots$ \\

\midrule
$\vdots$  &$\vdots$ &$\vdots$ &$\vdots$ &$\vdots$ &$\vdots$ &$\vdots$ \\
\midrule
Erasure ($\epsilon_M = 0.9$) &X &X & &X &X &$\cdots$ \\
\textbf{Agent M} ($\Tilde{a}_{t}^{(M)}$) &$\Tilde{a}_{0}^{(M)}$ &$\Tilde{a}_{0}^{(M)}$ &$a_3^{(M)}$ &$a_3^{(M)}$ &$a_3^{(M)}$ &$\cdots$ \\
\bottomrule
\end{tabular}
\begin{tablenotes}
\item X denotes the erasure of the action for the given round and agent.
\end{tablenotes}
\end{threeparttable}
\end{table*}

\subsection{Multi-Agent Multi-Armed Bandits with Action Erasures} \label{mamab_setting} Consider a central learner connected to $M$ distributed agents, indexed by $[M]$, over heterogeneous erasure channels. The learner faces a stochastic $K$-armed bandit problem, i.e., $|\gA| = K$. At each round $t$ during a time horizon $T$, the learner selects an action $a_{t}^{(m)} \in \gA$ for each agent $m \in [M]$ to play. That is, $M$ actions are played per round (unlike the traditional setting described above). When an action is chosen for agent $m$, it is communicated through an i.i.d. action erasure channel characterized by erasure probability $\epsilon_m$, and may or may not be received. 
That is, independently from other rounds and agents, each agent $m$ receives $a_{t}^{(m)}$ with probability $1-\epsilon_m$ and does not receive an action with probability $\epsilon_m$. The learner does not know which action requests get erased, but has knowledge of upper bounds on the erasure probabilities. The agents, on the other hand, perceive their own erasures but they do not have a feedback mechanism to inform the learner, i.e., there is no uplink between the agents and the learner. Furthermore, as motivated from  applications discussed in \Secref{intro}, we assume that the agents 
cannot (or do not wish to) run the algorithm themselves 
and continues to play
the same action (last successfully received action), denoted as $ \Tilde{a}_{t}^{(m)} \in \gA$ 
to play in the case of an erasure. $\Tilde{a}_{0}^{(m)} \in \gA$ denotes the action performed by the agent $m$ if the action in first round is erased, it is chosen uniformly at random. An example of multi-agent multi-armed bandit learning with action erasures is provided in Table~\ref{table:system-model-example}.

{\bf Observation.} {Although we focus on channels with erasures, our model can also apply over action delays: an agent receives an action not at the timeslot sent, but at a later time, based on a (known) delay probability distribution, and only changes the action she plays once she receives a new action. Our algorithms naturally extend and apply to this setting as well.}

{\bf Design Objective.} Our objective is to formulate a distributed learning policy composed of two key elements: a decision strategy that directs the selection of actions $a_{t}^{(m)}$ for each agent $m$ at each time $t$, and a coping mechanism for the possible mismatch between the selected actions and received rewards due to erasures. The performance metric we want to optimize is the total cumulative regret incurred by the policy over time $T$ and over all $M$ agents:
\[
    R_T = \sum_{m=1}^{M} \left( T  \max_{a \in \gA} \mu_a - \mathbb{E}\left[ \sum_{t=1}^{T} r^{(m)}_t \right] \right)
\]

We note that the cumulative regret in a perfect communication setting (no action erasures) is lower bounded by $\Omega(\sqrt{KMT})$. This bound corresponds to the optimal regret order in a centralized $K$-armed bandit setup, where a total of $MT$ reward observations are centrally accessible for learning.


\section{PROPOSED ALGORITHM}
\label{algorithm}

In this section, we introduce Batched Scheduled Persistent Pulls (\algoname), a Successive Arm Elimination (SAE) 
based multi-agent multi-armed bandit algorithm with a crafted scheduling part. The pseudocode can be found in \Algref{main}.

For the problem we consider, misinformation (associating rewards with the wrong action)  can create shifts in the action means. For instance, in the erasures example in Table \ref{table:system-model-example}, at time $t=3$ the learner observes the reward of the action agent $M$ plays, but does not know whether this reward is associated with action $\Tilde{a}_{0}^{(M)}$, $a_1^{(M)}, a_2^{(M)}$, or $a_3^{(M)} .$
Intuitively, to minimize this shift, it is meaningful to study an algorithm where the same action pulls are repeated several times; in the example in Table~\ref{table:system-model-example}, if we had selected $a_1^{(M)}=a_2^{(M)}=a_3^{(M)}=a$,
then, we could correctly associate reward at time $t=3$ with action $a$. Moreover,
the fact that we need to play in parallel across $M$ agents, implies that we need to use a batched algorithm. Accordingly, we base our proposed algorithm on SAE \ifarxivFormat\cite{auer2010ucb} \else\citep{auer2010ucb}\fi, described next, with modifications that enable  robustness to misinformation.

SAE is a batched algorithm, i.e., it divides the horizon into batches of exponentially increasing length and eliminates actions based on a shrinking confidence region defined by the number of pulls \ifarxivFormat\cite{auer2010ucb} \else\citep{auer2010ucb}\fi. In each batch $i$, all remaining actions, included in a set $\gA_i$, are pulled $4^i$ times, and after all pulls of the batch are completed, actions are retained if:
\begin{equation}\nonumber
    \gA_{i+1} \gets \{ a \in \gA_i | \max\limits_{\Bar{a} \in \gA_i} \hat{\mu}_{\Bar{a}}^{(i)} - \hat{\mu}_{a}^{(i)} \leq 4 \sqrt{\log{(KT)} / 2 \cdot 4^i}
\}  \end{equation} where $\hat{\mu}_{a}^{(i)}$ indicates the empirical mean of the reward of action $a$ in batch $i$.

Note that applying SAE directly 
in our setup does not perform well, due to two issues that need attention: (1) it may eliminate the best arm in early batches due to wrong feedback resulting in linear regret, (2) allocate an unnecessary amount of resource to bad channels.
We have to modify SAE to address these two issues, as otherwise, as Examples 1 and 2 later in this section illustrate, we may accrue large regret.

Addressing the first issue is straightforward:
the learner  simply repeats each action\footnote{A similar scheme was proposed in \ifarxivFormat\cite{hanna2023multi} \else\citet{hanna2023multi} \fi for the case of a single agent system.}
until the probability the correct action has been successfully received by the agent is sufficiently high. Only after this point the learner starts associating rewards with actions, thus minimizing the probability of misinformation. 
More specifically, if the learner decides to receive $p$ number of rewards for an action $a$ through agent $m$, the  learner  first asks the agent $m$ to repeat the action $\alpha_m = \lceil 4 \log{T} / \log{(1/\epsilon_m)} \rceil - 1$ additional times to ensure a success probability of at least $1-1/poly(T)$. A total of $\alpha_m + p$ rewards are generated in the environment, but only the last $p$ are taken into account by the learner to update the mean estimate of action $a$. This ensures with high probability that the rewards considered (effective rewards) are generated from the distribution associated with the selected action. We note however that all $\alpha_m + p$ rewards generated are counted in our regret, and thus, large $\alpha_m$ values  can affect the regret values we get, as we will  also see in Section~\ref{sec:analysis}.

Addressing the second issue, scheduling how action pulls are allocated across agents, is significantly more challenging. 
One issue is that, we need to wait for all pulls of batch $i$ to finish (agents that finish their tasks earlier will simply play random actions, potentially accumulating regret) before starting the next batch. Thus, the total regret we will accrue at batch $i$, is mainly determined\footnote{Recall that all actions in the set $\gA_i$ are expected to have mean values within a bounded distance from the optimal; thus the suboptimal actions in $\gA_i$ are expected to accumulate similar regret.}
by  $ T^{(i)}$, 
the time at which all $4^i$ pulls of the actions in $\gA_i$ are completed; and $ T^{(i)}$ 
 highly depends on the schedule, as simple examples can illustrate.



The following examples illustrate that two  (natural to consider) scheduling algorithms (one playing all $4^i$ pulls of an action at only one agent, and the other splitting the pulls of each action across all agents) can lead to larger than needed $T^{(i)}$ and thus suboptimal regret.


\textbf{Example 1} \label{para:example-vertical} 
Assume that we order the agents so that $\alpha_1\leq \alpha_2\leq \ldots \alpha_M$ (where $\alpha_m$ is the number of repetitions in each channel to ensure high probability action delivery). One intuitive schedule could be, to assign $\lfloor K^{(i)} / M \rfloor$ actions to each agent and place the remaining $\hat{K} = K - \lfloor K^{(i)} / M \rfloor M$ actions to the first (fastest) $\hat{K}$ agents, where $K^{(i)}$ is the number of active actions in batch $i$, i.e., $|\gA_i| = K^{(i)}$ . This scheduling has end time $T^{(i)} = \max{\left( \lfloor \frac{K^{(i)}}{M} \rfloor (\alpha_M + 4^i), (\lfloor \frac{K^{(i)}}{M} \rfloor + 1)(\alpha_{\hat{K}} + 4^i) \right)}$ - and although for some $\alpha_i$ values it can perform well, it also fails in many scenarios. For instance, if $\epsilon_m = 0 ~\forall m \in [M]$, the end time becomes $4^i K^{(i)}$ whereas the optimal end time is $\lceil 4^i K^{(i)}/M \rceil$ (which is smaller by a factor of $M$). 

\textbf{Example 2} \label{para:example-horizontal} Another straightforward approach is to first complete $4^i$ pulls for one action by distributing the pulls across all agents, and then move onto the next. That is, for each action, the learner sends it to all agents, and waits until $4^i$ effective pulls (i.e., not counting repetitions) are received back. Note that even if an agent $m$ needs to play one pull, we still need to wait first for $\alpha_m$ rounds before collecting this effective reward. This scheduling has an end time $ T^{(i)} = K^{(i)} \min\limits_{\Tilde{M} \in [M]} (\frac{\sum_{m=1}^{\Tilde{M}} \alpha_m + 4^i}{\Tilde{M}}) $,  where $K^{(i)}=|\gA_i|$. 
This can be suboptimal, e.g., if $\alpha_m=\alpha ~\forall m \in [M]$, then the end time of this scheduling is $T^{(i)} = K^{(i)} \alpha + K^{(i)} \lceil \frac{4^i}{M} \rceil$, whereas an end time of $\lceil \frac{K^{(i)}}{M} \rceil (\alpha + 4^i)$ can be achieved using the scheduling explained in Example 1. 

\begin{algorithm}
    \caption{\label{main} \algoname ($K$, $M$, $\valpha$)}
    \begin{algorithmic}[1]
    \State{\textbf{Input:} number of actions $K$, number of agents $M$, repetitions $\valpha \in \sZ^M_{+}$}
    \State{Initialize batch index $i = 1$, set of active actions $\gA_1 = \left[ K\right]$ }
    \For{batch $i$}
        \State{$ \mS, T^{(i)}$ = Schedule$(\gA_i, \valpha, i )$ (see \Algref{scheduling-alg})}
        \For{$t$ in $[T^{(i)}]$} 
            \State{send action $\emS_{mt}$ to agent $m$ $\quad \forall m \in \left[ M \right]$}
            \State{receive reward $r_{mt}$ $\quad \forall m \in \left[ M \right]$}
        \EndFor
        \State{Update means of the actions \[~~\evmu^{(i)}_a = \sum\limits_{m \in \sM^{(i)}_a} \sum_{t = b_{am}^{(i)}}^{e_{am}^{(i)}} r_{mt} / 4^i \quad \forall a \in \gA_i \] \label{step:emp-mean}\\ $\quad $  ($\sM^{(i)}_a$: set of agents that pulls action $a$ in batch $i$, $b_{am}^{(i)}$ and $e_{am}^{(i)}$: start and end time of the effective pulls, respectively, of $a$ in agent $m$ in batch $i$, $b_{am}^{(i)}, e_{am}^{(i)}  \in [T^{(i)}]$) }
        \State{Update active action set: $\gA_{i+1} \gets \{ a \in \gA_i | \max\limits_{j \in \gA_i } \evmu^{(i)}_j - \evmu^{(i)}_a \leq 4 \sqrt{\log{(KMT)} / 2 \cdot 4^i} \} $ }
        \State{$i \gets i + 1$ }
    \EndFor
    \end{algorithmic}
\end{algorithm}



Our scheduling goal is, given $\alpha_1\leq \alpha_2\leq \ldots \leq\alpha_M$, to find a schedule that minimizes $T^{(i)}$.
As the previous examples illustrate, neither distributing actions across all agents, nor restricting each action to be played in one agent, is 
optimal.
One natural approach is to express the schedule through an Integer Linear Program  (provided in appendix \ref{app:lp-formulation}). The associated LP relaxation, as also discussed in the appendix, essentially associates a cost $\frac{a_m}{4^i}$ with each action pull at agent $m$, and solves a cost-minimization resource allocation problem. 
The resulting LP solution gives us a lower bound $\tau$ on $T^{(i)}$, where:
\begin{equation}
\tau := 4^i K^{(i)} / \sum_{m=1}^{M} 1 / (\frac{\alpha_m}{4^i} + 1). \label{tau}
\end{equation}
Unfortunately, the LP solution cannot always be easily translated to an integral solution (where each agent $m$ actually plays $a_m$ pulls even if she needs to collect reward for an action only once) while avoiding suboptimal regrets (as compared to the ILP solution).

Instead, we develop a scheduling algorithm that is polynomial time, and carefully balances how 
to split the $4^i$ pulls of each action  across agents,
so as to decrease the number of required repetitions $\alpha_i$, while still taking advantage as needed from the fact that we have multiple agents.
  The pseudocode can be found in Algorithm~\ref{scheduling-alg}.

{The algorithm works in two stages: We first round the LP solution to an integer solution, which can schedule at least  $(K-M)^+$ actions. In this stage, each action is assigned to at most one agent. In the second stage, we schedule the remaining unscheduled actions by splitting each action among multiple agents.
In particular, we assign to the first stage (where we do not split action pulls)  a duration $\tau$ as in (\ref{tau}): since the LP relaxation manages to allocate the $4^i$ pulls for all actions before $\tau$, keeping all pulls of an action together before that time can only decrease the total number of repetitions required by each allocated action. We prove in Section~\ref{sec:analysis} that at least  $(K-M)^+$\footnote{$x^+ = \max{(x, 0)} ~~\forall x \in \mathbb{R}$} actions will be successfully allocated at this stage, leaving $\hat{K}$ remaining actions.
In the second stage, we partition the pulls of the remaining  $\hat{K}$ actions into smaller parts of size $\max{(1, \lfloor M / 2\hat{K}\rfloor)}$ and use the first $\lfloor M / 2 \rfloor$ agents to do the scheduling. Utilizing only the first $\lfloor M / 2 \rfloor$ agents allows to find an end time on the scheduling in terms of $c \sum_{m=1}^{M} \alpha_m$ where $c > 0$ is some constant instead of a term that depends on $K$ or $M$, as will become apparent in \Secref{sec:analysis}.}

\begin{algorithm}
    \caption{\label{scheduling-alg} Schedule ($\gA$, $\valpha$, $i$)}
    \begin{algorithmic}[1]
    \State {\textbf{Input:} set of actions $\gA$ with $|\gA| = K$, repetitions $\valpha \in \sZ^M_{+}$, batch index $i$}
    \State Initialize $k= 0,~ T^{(i)} = \tau$ (see  Eq.~\ref{tau})
    \State{Shuffle the set $\gA$ randomly}
    \For { agent $m \in [M]$}
        \State {Initialize $t_{end} = 0,~ p = \evalpha_m + 4^i$ }
        \While {$t_{end} + p \leq T^{(i)}$}
            \State {Assign next action to agent}
            \State {$k \gets k + 1$}
            \State {$t_{\text{end}} \gets t_{\text{end}} + p$}
        \EndWhile
    \EndFor
    \For {$\hat{K} = K - k$ unassigned actions} \label{schedalg:Khat}
        \State {Divide pulls into $\max{(1, \lfloor M / 2\hat{K}\rfloor)}$ equal parts}
    \EndFor
    \State {Assign each part to first $\lfloor M / 2 \rfloor$ agents one by one}
    \State {Imitate assignments of first $\lfloor M / 2 \rfloor$ agents for remaining  $M - \lfloor M / 2 \rfloor$ agents (with their own repetitions)}
    \State {Update $T^{(i)}$, the end time of the batch, to agent finishing last in first $\lfloor M / 2 \rfloor$ agents}
    \State {Fill remaining slots of the agents randomly}
    \State {\textbf{Output:} $\mS \in \sR^{M \times T^{(i)}}$ the schedule of actions to agents, $T^{(i)}$ end time }
    \end{algorithmic}
\end{algorithm}

\subsection{Connecting to Channels with Delays}
\label{sec:delay_interpretation}
We note that our algorithm BatchSP2 (and its analysis, in the next section),
directly applies to channels with delays, where an action sent by the learner to an agent $m$ is received after $t$ rounds with some probability $p_t^{(m)}$. Indeed, although we used erasure channels for our narrative in this paper, and motivated use of repetitions over such channels, the only fact that BatchSP2 essentially hinges on is that, 
 agent $m$ will receive a sent action with probability at least $\frac{1}{T}$ after $\alpha_m$ rounds, where $\alpha_m$ is known. \violet{It implies that any known (or estimated) delay/probability of successful reception can be used with our algorithms.}  In our case $\alpha_m$ was dictated from the repetition protocol, in other setups it could be dictated from delivery delay or delivery uncertainty. \violet{Datasets and models in literature, e.g., \ifarxivFormat\cite{datasetonewaydelay2020} or \cite{dahmouni2012analyticaljitter}\else\citep{datasetonewaydelay2020} or \citep{dahmouni2012analyticaljitter}\fi, can provide empirical values for delay/probability in such setups.}


\section{REGRET ANALYSIS}
\label{sec:analysis}
This section provides our theoretical analysis: we first calculate an upper bound on the end time of each batch in Lemma~\ref{lem:schedule-endtime}, then use this to derive an upper bound on the expected regret that depends on suboptimality gaps on \Theoref{main-thm}, and provide a gap-independent regret upper bound on \Theoref{instance-indep-thm}.

\begin{restatable}{lemma}{lemmasched}
    \label{lem:schedule-endtime}
    If the scheduling algorithm outlined in \Algref{scheduling-alg} is run for batch $i$, then the end time $T^{(i)}$ of the batch can be bounded as
    \[
        T^{(i)} \leq K 4^i \tau + 6 \left( \sum_{m=1}^{M} \frac{\alpha_m}{M} + 2 \frac{K 4^i}{M} \right)
    \] where $\tau = \frac{1}{ \sum\limits_{m=1}^{M} 1 / (\alpha_m / 4^i + 1)}$, $\alpha_m = \lceil 4 \frac{\log{T}}{ \log{(1/\epsilon_m)}}\rceil -1$, $K$ is the number of actions, and $M$ is the number of agents.
\end{restatable}
\noindent{\textit{Proof Sketch of Lemma~\ref{lem:schedule-endtime}.}
The upper bound on the scheduling end time, hence, total number of pulled actions in a batch, is obtained in two steps. First, because  $\alpha_m+4^i$ rounds are sufficient to schedule $4^i$ effective pulls for a single arm at agent $m$, we prove that Algorithm~\ref{scheduling-alg} schedules at least $(K-M)^+$ agents in time $K4^i \tau$. This implies that at step $10$ of Algorithm~\ref{scheduling-alg}, the number of remaining arms to be scheduled is bounded by $M$. As the algorithm schedules these arms among the best $M/2$ agents, each agent will be assigned a constant number of arms. The final end time is bounded by noticing that from the averaging principle, the delay of all agents in the best half is bounded by the average delay $\sum_m \alpha_m/M$.

The complete proof is provided in Appendix \ref{app:proof-lemma-1}.
\ifarxivFormat\else\newpage\fi
\begin{restatable}{theorem}{theoremone}
    \label{main-thm}
    Consider a distributed multi-armed bandit setting  with $K$ actions and $M$ agents connected through heterogeneous erasure channels with erasure probabilities $\{ \epsilon_i\}_{i=1}^{M}$. If \algoname\ is run with horizon $T$, then the expected regret is,
    \ifarxivFormat
    \begin{align*}
        \mathbb{E}[R_T] \leq c \Bigg( \Bigg. \sum_{a: \Delta_a > 0} \Big( \Big.  \frac{\log{(KMT)}}{\Delta_a} + \frac{M \log{(MT)}}{\sum\limits_{m=1}^{M} 1 / (\alpha_m + \frac{\log{(KMT)}}{\Delta_{a}})} \Big. \Big) + \sum_{m=1}^{M} \alpha_m \log{(MT)}  +  \log{(MT)} \Bigg. \Bigg)
    \end{align*} 
    \else
    \begin{align*}
        \mathbb{E}[R_T] \leq c \Bigg( \Bigg. &\sum_{a: \Delta_a > 0} \Big( \Big.  \frac{\log{(KMT)}}{\Delta_a} \\ 
        &~~~~ + \frac{M \log{(MT)}}{\sum\limits_{m=1}^{M} 1 / (\alpha_m + \frac{\log{(KMT)}}{\Delta_{a}})} \Big. \Big) \\ + & \sum_{m=1}^{M} \alpha_m \log{(MT)}  +  \log{(MT)} \Bigg. \Bigg)
    \end{align*}
    \fi
   where $\alpha_m = \lceil 4 \log{T} / \log{(1/\epsilon_m)}\rceil -1$ is the number of repetitions at agent $m$, $\Delta_a$ is the suboptimality gap for action $a$, and $c > 0$ a constant.
\end{restatable}

\noindent{\textit{Proof Sketch of \Theoref{main-thm}.} The regret bound is achieved by decomposing the regret of each batch as $\mathbb{E}[R_T^{(i)}]= \sum_a T_{ia} \Delta_a$ and bounding the expected number of times $\mathbb{E}[T_{ia}]$ that arm $a$ is pulled in batch $i$. To that end, we condition on a good event entailing that for each agent $m$ and each consecutive $\alpha_m+x$ pulls from action $a$, the last $x$ rewards are samples from the distribution of action $a$. This provides a concentration of the empirical means used in Algorithm~\violet{\ref{main}} with high probability. As a result, we get an upper bound on the number of batches a suboptimal arm can survive. Having this, to bound $T_{ia}$, it only remains to bound the number of times an active action is pulled in batch $i$, \violet{which is highly sensitive to the scheduling of action pulls}. This is proved by utilizing the upper bound on the number of pulls in Lemma~\ref{lem:schedule-endtime} and \violet{leveraging} the symmetry imposed by the randomization in Algorithm~\ref{scheduling-alg} to show that each action has an equal contribution in the total number of pulls. The final regret bound is obtained by showing that the good event has high probability.

Bounding the excess regret from the rewards not used by the algorithm is a challenging part of the regret analysis. If the schedule is designed naively, these rewards may come from the action with the largest gap in the batch. However, as we show in the proof of Theorem~\ref{main-thm}, the randomization and shuffling performed in Algorithm~\ref{scheduling-alg} make the contributions of the different arms in the excess regret uniform in expectation.

The complete proof is provided in Appendix \ref{app:proof-thm-1}.

\violet{The three components of the regret bound in Theorem~\ref{main-thm} originate from distinct aspects of the algorithm. The initial term, $\sum_{a: \Delta_a > 0} \log{(KMT)} / \Delta_a$, is inevitable, representing the order optimal regret achievable under perfect channels (no delay, no erasure).
The second and third terms are due to the repetition and scheduling of actions (Algorithm~\ref{scheduling-alg}). It is noteworthy that, the second term matches the regret of an optimal scheduling algorithm (see App.~\ref{app:lp-formulation}).
Additionally, under perfect channels, this term simplifies to the a lower bound on the regret up to logarithmic factors.
The third term, $\sum_{m=1}^{M} \alpha_m \log{(MT)}$, emerges at the outset of the learning process, reflecting that each agent $m$ will repeat the first pulled (suboptimal) action for $\alpha_m$ iterations on average.}

The regret bound in Theorem~\ref{main-thm} is nearly constant for constant gaps and erasure probabilities. However, for small gaps, the regret bound can be large. It is important to note that this will not be the actual regret suffered when the gaps are small, as for small $\Delta$, the regret is bounded by $TM\Delta$. The following theorem provides an instance-dependent regret bound that works for all values of the suboptimality gaps.

\begin{restatable}{theorem}{theoremtwo}
    \label{instance-indep-thm}
    Consider the distributed multi-armed bandit setting with $K$ actions and $M$ agents connected through heterogeneous erasure channels $\{ \epsilon_i\}_{i=1}^{M}$. If \algoname is run for horizon $T$, then the expected regret is
    \ifarxivFormat
    \begin{align*}
        R_T \leq c \Bigg( \Bigg. M \sqrt{\frac{ K T \log{(MT)} }{\sum_{m=1}^{M} 1 / (\alpha_m\Delta_{\star} + \log{(KMT)})}} + \sum_{m=1}^{M} \alpha_m \log{(KMT)}\Bigg. \Bigg)
    \end{align*}
    \else
    \begin{align*}
        R_T \leq c \Bigg( \Bigg. & M \sqrt{\frac{ K T \log{(MT)} }{\sum_{m=1}^{M} 1 / (\alpha_m\Delta_{\star} + \log{(KMT)})}} \\ &~ + \sum_{m=1}^{M} \alpha_m \log{(KMT)}\Bigg. \Bigg)
    \end{align*} \fi     
   where $\Delta_{\star}$ is the value satisfying \[
    \Delta_{\star} = \frac{c' K \log{(MT)}}{T \sum\limits_{m=1}^{M} 1 / (\alpha_m + \frac{\log{(KMT)}}{\Delta_{\star}})},
   \] which can be efficiently approximated using the bisection method, $\alpha_m = \lceil 4 \log{T} / \log{(1/\epsilon_m)}\rceil-1$ number of repetitions and $c, c' > 0$ constants.
\end{restatable}

\noindent{\textit{Proof Sketch of \Theoref{instance-indep-thm}.} The regret bound is proved by bounding the regret bound for arms with small gap (less than $\Delta_\star$\violet{, which will be determined later}) by $TM\Delta_\star$ and using the bound in Theorem~\ref{main-thm} for the remaining arms with large gap (greater than $\Delta_\star$). The value of $\Delta_\star$ \violet{is chosen to minimize the bound by} balancing the regret resulting from arms with small gaps and the regret from arms with large gaps.


The complete proof is provided in Appendix \ref{app:proof-thm-2}.


It is worth noting that for $\alpha_m=0 \quad \forall m$ (no erasures), the regret bound in Theorem~\ref{instance-indep-thm} reduces to $\tilde{O}(\sqrt{KTM})$, nearly matching the lower bound on the regret for the model considered in \ifarxivFormat\cite{lattimore2020bandit} \else\citet{lattimore2020bandit}\fi. More importantly, our bound shows that if $\alpha_m=\tilde{O}(1/\Delta_{\max}) \quad \forall m$, where $\Delta_{\max}$ is the maximum gap, then the regret bound is still $\tilde{O}(\sqrt{KTM})$, hence, we (nearly) suffer no extra regret beyond the no erasure case.

For the single agent case $M=1$, the regret bound in Theorem~\ref{instance-indep-thm} reduces to $\tilde{O}(\sqrt{KT}+K\alpha)$ which is shown to be nearly optimal in \ifarxivFormat\cite{hanna2023multi} \else\citet{hanna2023multi}\fi.

\section{EXPERIMENTS}
\label{experiments}

\begin{figure*}[tbh!]
  \centering
  \ifarxivFormat\includegraphics[width=1\linewidth]{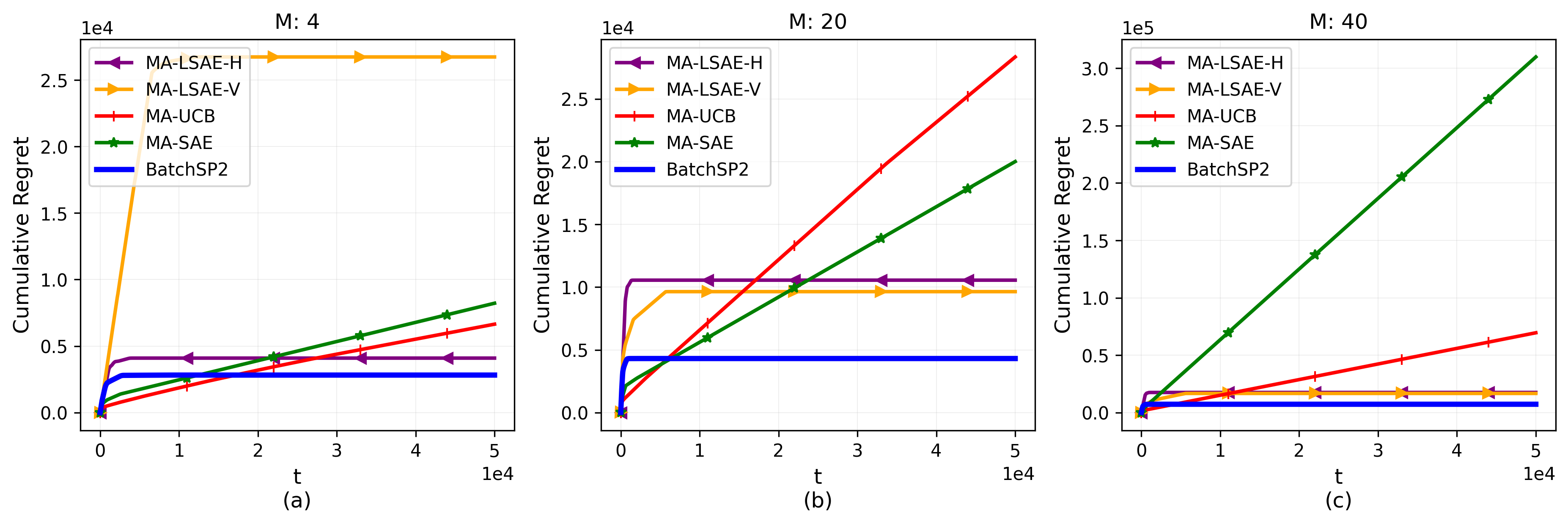}
  \else\includegraphics[width=0.95\linewidth]{figures/k10_m4-20-40_marked.png}\fi
  \ifarxivFormat
    \captionsetup{justification=centering}
  \fi
  \caption{Comparison Results For Different Numbers Of Agents. From Left To Right, The Plots Show Cumulative Regret As A Function Of Rounds t For (a) 4 Agents, (b) 20 Agents, and (c) 40 Agents, Respectively.}
  \label{fig:k10_m4-20-40}
\end{figure*}

In this section, we empirically evaluate the regret performance of our proposed algorithm, BatchSP2, and compare against the following methods:
\begin{itemize}
    \item MA-SAE: This is an extension of SAE \ifarxivFormat\cite{auer2010ucb} \else\citep{auer2010ucb} \fi to multi-agent setting. It utilizes the agents without repeating any actions and considers all the rewards generated in the environment. 
    \item MA-LSAE-V: This is an extension of SAE to multi-agent setting, that restricts all pulls of an action to be played at the same agent, as described in Example~1 in \Secref{algorithm}. 
    
    \item MA-LSAE-H: This is another extension of SAE to multi-agent setting, that distributes the pulls of an action across all agents, as described in Example~2 in \Secref{algorithm}. 
    \item MA-UCB: This is an extension of Upper Confidence Bound (UCB) \ifarxivFormat\cite{Auer2002b, Lai1987AdaptiveTA} \else\citep{Auer2002b, Lai1987AdaptiveTA} \fi algorithm to multi-agent setting. UCB is an optimal algorithm for a simple MAB setting. Compared to SAE, it makes the decision on which action to pull at each round instead of at each batch.
\end{itemize}

We have explored a number of experimental setups (in terms of number of actions, channel quality, horizons, etc)\footnote{\violet{The code to our experiments is available \href{https://github.com/mervekarakas/mamab_erasures/}{here}}.}; we here show results for two experiments that we believe are representative:

$\bullet$ {\bf Experiment 1}, shown in Figure~\ref{fig:k10_m4-20-40}, uses $K=10$ actions, with Gaussian reward distributions that have variance $1$ and means $\left[0.8, 1, 0, \cdots, 0\right]$.  The time horizon is $T=5 \times 10^4$ and the regret in each plot is averaged over $100$ experiments with arms shuffled. The channels have erasure probabilities  $0.2$, $0.7$, $0.9$, and $0.99$, and there is an equal number of $M/4$ channels for each erasure probability.

$\bullet$ {\bf Experiment 2}, shown in Figure \ref{fig:higher_erasures}, has all parameters the same as Experiment 1, with the difference that we have now channels with similar erasure probabilities of $0.9$, $0.93$, $0.95$, and $0.99$ (as before, there is an equal number of channels for each erasure probability).

From Figure~\ref{fig:k10_m4-20-40}, it can be seen that extensions of UCB and SAE may result in linear regret under action erasures even when the suboptimality gap is large (0.2 for the instance used in this experiment). Comparing Figure~\ref{fig:k10_m4-20-40} (a) to (c) for MA-LSAE-V, we can observe how waiting for a bad channel to finish pulling actions slows down the learning: when the number of agents with erasure probability 0.2 increases from left to right, MA-LSAE-V starts assigning actions to agents with small number of repetitions, and cumulative regret gets smaller. This supports the splitting idea behind our algorithm. 

From Figure~\ref{fig:higher_erasures}, it can be seen that while trends of algorithms are similar to Figure~\ref{fig:k10_m4-20-40} (b) in terms of learning (linear versus logarithmic cumulative regret); when the channel quality gets worse, the gap between MA-LSAE-H and our algorithm widens. This indicates that while repetitions ensure learning with high probability, if we assign action pulls to agents without considering how many additional repetitions are evoked, it can significantly slow down the learning process. In some cases, it might even result in UCB or SAE to having lower regret for an extended period of time, despite their linear regret behavior.

\begin{figure}[t!]
  \centering
  \ifarxivFormat\includegraphics[width=0.45\textwidth]{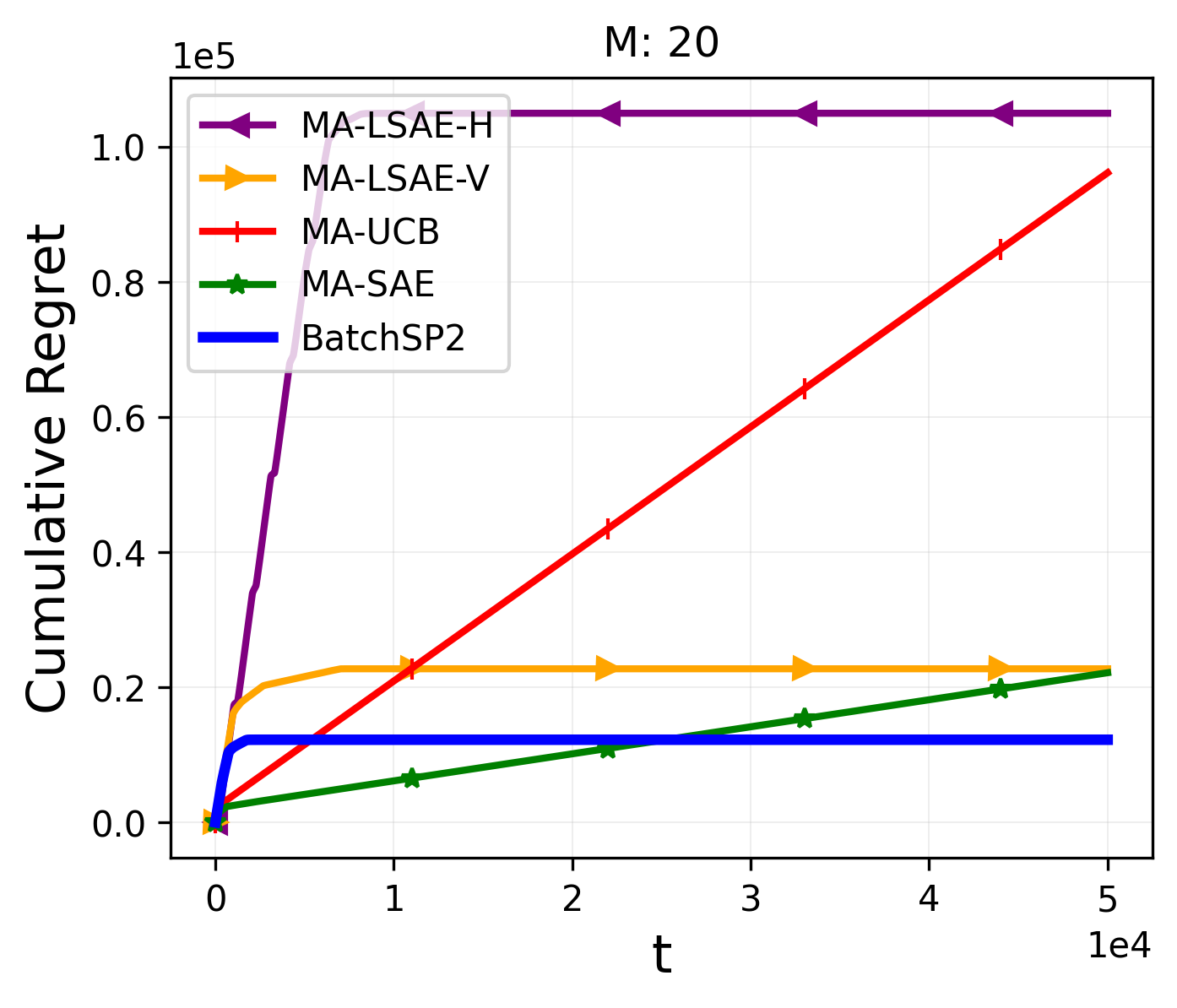}\else
    \includegraphics[width=0.35\textwidth]{figures/k10_m20_high_erasures_marked.png}
  \fi
  \ifarxivFormat
  \captionsetup{justification=centering}
  \fi
  \caption{Same Scenario As In Figure~\ref{fig:k10_m4-20-40} (b) With Worse Channel Quality.}
  \label{fig:higher_erasures}
\end{figure}

\section{CONCLUSION}
\label{conclusion}
In this work, we consider the case of a learner connected to multiple distributed agents through heterogeneous channels, that are subject to action erasures without action feedback (the same setup can also capture delays and uncertainty on action reception). If rewards can be externally observed, we may have misinformation, a mismatch between the action the learner requests and the agent plays. Because of this, traditional algorithms can easily fail; instead, we introduce BatchSP2, an efficient algorithm that uses repetition to achieve robustness over erasures and careful allocation of action pulls to agents to minimize regret.  We provide  a theoretical regret analysis of  BatchSP2, which allows to recover as special cases existing bounds,
as well as numerical evaluations that show Batch2SP can achieve superior performance over baseline schemes.

\subsubsection*{Acknowledgements}
\violet{We thank the anonymous reviewers and the meta-reviewer for their insightful suggestions and comments. This work is supported in part by NSF grants \#2007714 and \#2221871, by Army Research Laboratory grant under Cooperative Agreement W911NF-17-2-0196, and by the Amazon Faculty Award.}

\appendices
\newpage
\section{LINEAR PROGRAM FORMULATION}
\label{app:lp-formulation}
We first formulate a (nonlinear integer) program  that minimizes the end time to schedule action pulls in batch $i$ with $K$ actions across $M$ agents as follows:
\begin{subequations}
\label{p:formulation}
\begin{empheq}[box=\widefbox]{align}
    \min\limits_{\mX \in \mathbb{R}^{M \times K} } &~~ \max\limits_{m \in [M]} \sum\limits_{k=1}^{K} \left( \alpha_m \mathds{1}[X_{mk} > 0] + X_{mk} \right) \label{p:objective}\\
    \text{s.t.} ~~&~~  \sum\limits_{m=1}^{M} X_{mk} = 4^i ~~~\forall k \in [K] \label{p:sum}\\
    &~~  X_{mk} \in \{0,1,2,\cdots,4^i \} ~~~\forall m \in [M], ~\forall k \in [K] \label{p:integer},
\end{empheq}
\end{subequations}where $\mX \in \mathbb{R}^{M \times K}$ captures the variables of the program, with $X_{mk}$ indicating the number of effective pulls of action $k$ performed by agent $m$. The objective function (\ref{p:objective}) is to minimize the latest end time among agents. Constraint ($\ref{p:sum}$) ensures that the total number of effective pulls for each action is $4^i$; and constraint (\ref{p:integer}) forces effective pulls assigned to each agent per action to be an integral value in $[0,4^i]$. It is easy to see that the progam in (\ref{p:formulation}) is equivalent to the following  integer linear program (ILP): 
\begin{subequations}
\label{ilp:formulation}
\begin{empheq}[box=\widefbox]{align}
    \min\limits_{\substack{\mX, \mW \in \mathbb{R}^{M \times K} \\  t \in \mathbb{R} }} &~~~~ t \label{ilp:objective}\\
    \text{s.t.} ~~~~~&~~ \sum\limits_{k=1}^{K} \left( \alpha_m W_{mk} + X_{mk} \right) \leq t,  ~~~\forall m \in [M] \label{ilp:obj_to_const}\\
    &~~ X_{mk} \leq 4^i W_{mk} ~~~\forall m \in [M], ~\forall k \in [K] \label{ilp:indicator}\\
    &~~  \sum\limits_{m=1}^{M} X_{mk} = 4^i ~~~\forall k \in [K] \label{ilp:sum}\\
    &~~  X_{mk} \in \{0,1,2,\cdots,4^i \} ~~~\forall m \in [M], ~\forall k \in [K] \label{ilp:integer_x} \\
    &~~  W_{mk} \in \{0,1 \} ~~~\forall m \in [M], ~\forall k \in [K] \label{ilp:integer_w},
\end{empheq}
\end{subequations} where the variable $t \in \mathbb{R}$ replaces the $\max$ in objective \eqref{p:objective} and the variable $\mW \in \mathbb{R}^{M \times K}$ replaces the indicator function. Notice that for any feasible solution $\mX$, if $X_{mk} > 0$, $W_{mk} = 1$. The relaxed version of the ILP in (\ref{ilp:formulation}) can be written as 

\begin{subequations}
\label{lp:relaxed_formulation}
\begin{empheq}[box=\widefbox]{align}
    \min\limits_{\substack{\mX, \mW \in \mathbb{R}^{M \times K} \\  t \in \mathbb{R} }} &~~~~ t \label{lp:relaxed_objective}\\
    \text{s.t.} ~~~~~&~~ \sum\limits_{k=1}^{K} \left( \alpha_m W_{mk} + X_{mk} \right) \leq t,  ~~~\forall m \in [M] \label{lp:relaxed_obj_to_const}\\
    &~~ \mX_{mk} \leq 4^i W_{mk} ~~~\forall m \in [M], ~\forall k \in [K] \label{lp:relaxed_indicator}\\
    &~~  \sum\limits_{m=1}^{M} X_{mk} = 4^i ~~~\forall k \in [K] \label{lp:relaxed_sum}\\
    &~~  0 \leq X_{mk} \leq 4^i ~~~\forall m \in [M], ~\forall k \in [K] \label{lp:relaxed_integer_x} \\
    &~~ 0 \leq W_{mk} \leq 1 ~~~\forall m \in [M], ~\forall k \in [K] \label{lp:relaxed_integer_w},
\end{empheq}
\end{subequations}

Notice that the minimum value $W_{mk}$ can take is $X_{mk} / 4^i$ due to (\ref{lp:relaxed_indicator}); hence, by replacing $W_{mk}$ with its minimum value, we get the following linear program which gives a lower bound on the ILP (\ref{ilp:formulation}):

\begin{subequations}
\label{lp:formulation}
\begin{empheq}[box=\widefbox]{align}
    \min\limits_{\substack{\mX \in \mathbb{R}^{M \times K} \\  t \in \mathbb{R} }} &~~~~ t \label{lp:objective}\\
    \text{s.t.} ~~~~~&~~ \sum\limits_{k=1}^{K} X_{mk} \left(  \frac{\alpha_m}{4^i}  + 1 \right) \leq t  ~~~\forall m \in [M] \label{lp:obj_to_const}\\
    &~~  \sum\limits_{m=1}^{M} X_{mk} = 4^i ~~~\forall k \in [K] \label{lp:sum}\\
    &~~  0 \leq X_{mk} ~~~\forall m \in [M], ~\forall k \in [K] \label{lp:nonneg_x},
\end{empheq}
\end{subequations}
In the linear program (\ref{lp:formulation}), $X_{mk}$ is the variable that indicates how many effective pulls are assigned to agent $m$ for action $k$. (\ref{lp:sum}) forces each action to be pulled $4^i$ effective times; however, instead of an integer number of pulls, each agent is allowed to perform nonnegative fractional pulls. Furthermore, (\ref{lp:obj_to_const}) indicates that for each agent $m$, one effective pull has a cost of $\frac{\alpha_m}{4^i} + 1$.

\begin{claim}\label{claim:optimal_soln_lp}
    The optimal objective value of (\ref{lp:formulation}) satisfies, $t^\star = \sum_{k=1}^{K} X_{mk}^\star (\alpha_m / 4^i + 1) ~\forall m \in M$, where ($t^\star, \mX^\star$) is the optimal solution of \ref{lp:formulation}.
\end{claim}

\noindent{\textit{Proof of Claim~\ref{claim:optimal_soln_lp}.}
First, we observe that at least one of the inequalities in (\ref{lp:obj_to_const}) holds with equality, otherwise the value of $t^\star$ can be decreased leading to a better objective. Define the set of indices $\gE_i := \{m \in [M] : \sum_{k=1}^{K} X_{mk}^\star (\alpha_m / 4^i + 1) = t^\star\}$.

Now, assume Claim~\ref{claim:optimal_soln_lp} is not correct. And let $m_s$ be such that$$\sum_{k=1}^{K} X_{m_sk}^\star (\alpha_{m_s} / 4^i + 1) < t^\star.$$
Then $\forall m \in \gE_i ~\exists \{ \beta_{mk} \}_{k=1}^{K} \geq 0: \sum_k \beta_{mk} > 0 $ small enough such that 
\ifarxivFormat
\begin{align}
    &X'_{mk} = \begin{cases}
        X^\star_{mk} - \beta_{mk}, & m \in \gE_i,~\forall k \in [K] \\
        X^\star_{m k} + \sum_{k=1}^{K} \beta_{mk} , &~ m_s = m \\
        X^\star_{mk} &~ \text{otherwise}
    \end{cases} \nonumber \\
    &t' = \max_{m} \{ \sum_{k=1}^{K} X_{mk}^{'} (\alpha_m / 4^i + 1) \} < t^\star \nonumber
\end{align}
\else
\begin{align}
    &X'_{mk} = \begin{cases}
        X^\star_{mk} - \beta_{mk}, & m \in \gE_i,~\forall k \in [K] \\
        X^\star_{m k} + \sum_{k=1}^{K} \beta_{mk} , &~ m_s = m \\
        X^\star_{mk} &~ \text{otherwise}
    \end{cases} \nonumber \\
    &t' = \max_{m} \{ \sum_{k=1}^{K} X_{mk}^{'} (\alpha_m / 4^i + 1) \} < t^\star \nonumber
\end{align}
\fi
forms a feasible solution in (\ref{lp:formulation}) with a smaller objective value $t' < t^\star$; hence, $(t^\star, \mX^\star)$ cannot be optimal. Then at the optimal solution $(t^\star, \mX^\star)$, $t^\star = \sum_{k=1}^{K} X_{mk}^\star (\alpha_m / 4^i + 1) ~\forall m \in M$.

Using Claim~\ref{claim:optimal_soln_lp} and the constraint (\ref{lp:sum});\ifarxivFormat\begin{align}\label{t_star}
    4^i K = \sum_{m=1}^{M} \sum_{k=1}^{K} X^\star_{mk} = \sum_{m=1}^{M} \frac{t^\star}{ (\alpha_m/4^i + 1)} ~=~ t^\star \sum_{m=1}^{M} \frac{1}{(\alpha_m/4^i + 1) } \nonumber \\
    ~\Rightarrow~~ t^\star ~=~ \frac{4^i K}{\sum_{m=1}^{M} 1/ (\alpha_m/4^i + 1)}
\end{align} \else
\begin{align}\label{t_star}
    4^i K = \sum_{m=1}^{M} \sum_{k=1}^{K} X^\star_{mk} = \sum_{m=1}^{M} \frac{t^\star}{ (\alpha_m/4^i + 1)} ~=~ t^\star \sum_{m=1}^{M} \frac{1}{(\alpha_m/4^i + 1) }
    ~~\Rightarrow~~ t^\star ~=~ \frac{4^i K}{\sum_{m=1}^{M} 1/ (\alpha_m/4^i + 1)}
\end{align}\fi which justifies  \eqref{tau}.

\paragraph{Observation} Note that the solution of the relaxed LP (\ref{lp:formulation}) can be directly used for scheduling of actions by adding $\max{(2\alpha_{M-1}, \alpha_m )}$ to the end time $t^\star$. Since the relaxation in general does not give a feasible solution for the ILP, we add $\max{(2\alpha_{M-1}, \alpha_M)}$ to the end time of the relaxed ILP to guarantee a feasible solution for the ILP.
As the additional time slots accumulate regret across all agents, this can result in $\Omega(M\alpha_M)$ additional regret which can be large for large $M$. Our algorithm improves the $M$ factor in $M\alpha_M$.

\section{MISSING PROOFS}
\label{app:missing-proofs}

\begin{table*}[tbhp!]
\centering
\caption{Notation} \label{table:notation}
\begin{tabular}{lcl}
\toprule
$4^i$& : & Number of effective pulls in batch $i$ for each active action \\
$\gA$& : & Set of actions, $|\gA| = K$  \\
$\gA_i$& : & Set of active actions in batch $i$, $|\gA_i| = K^{(i)}$\\
$\alpha_m$& : & $= \lceil 4 \log{T} / \log{(1/\epsilon_m)}\rceil-1$, number of repetitions for agent $m$\\
$\Delta_a$& : & $= \max_{a' \in \gA} \mu_{a'} - \mu_a$, suboptimality gap for action $a$\\
$G$& : & The event that at least one instruction among the times $t,t+1,\cdots,t+\alpha_m-1$ will not be\\
&  & erased for all agents $m$ and all times $t$\\
$G_i'$& : & $ = \left\{ | \evmu^{(j)}_a - \evmu_a| ~\leq~ 2 \sqrt{\frac{\log{(KMT)}}{2 \cdot 4^j}}  ~~\forall a \in \gA_j,~ j \in [i-1] \right\}$, the event that empirical means of\\ 
&  &active actions in batch $j$ ($\forall a \in A_j$) is in confidence region for all batches until batch $i$\\
$M$& : & Number of agents \\
$\mu_a$& : & Reward mean of action $a$\\
$\mu^{(i)}_a$& : & The empirical mean calculated for action $a$ at batch $i$ (as defined in step~\ref{step:emp-mean} in Algorithm~\ref{main})\\
$N_1^{(i)}$& : & $=\frac{M}{\sum\limits_{m=1}^{M} 1 / (\alpha_m + 4^i)}$, a term that appears in regret\\
$N_2^{(i)}$& : &$=12 \cdot 4^i$, a term that appears in regret \\
$R_T$& : & Regret of $K$ arm bandit over $M$ channels with horizon $T$\\
$R^{(i)}_T$& : & Regret of batch $i$ for $K$ arm bandit over $M$ channels with horizon $T$\\
$T^{(i)}$& : & Length of the scheduling outputted by \Algref{scheduling-alg} for batch $i$ \\
$T_i$& : & The total number of instructions played by all agents due to instructions sent in batch $i$ \\
$T_{ia}$& : & Number of times action $a$ is played by agents due to an instruction sent in batch $i$ \\
$\hat{K}$& : & Number of actions unassigned in the first part of the scheduling (as described in \Algref{scheduling-alg} line \ref{schedalg:Khat})\\
\bottomrule
\end{tabular}
\end{table*}

\subsection{Proof of Lemma \ref{lem:schedule-endtime}}
\label{app:proof-lemma-1}
In this section, we present the detailed proof of Lemma \ref{lem:schedule-endtime}. \\

\lemmasched*

We prove the upper bound on the end time in two steps.\\ 
\textbf{Step A.} First, we claim that the algorithm uses the first $4^i K \tau$ rounds to schedule all $4^i$ pulls of at least $(K-M)^{+}$ actions: \\ 
Each agent $m$ takes $\alpha_m + 4^i$ to complete all pulls of an action; hence, it can play all pulls of at least
\[
    \left\lfloor \frac{4^i K \tau}{\alpha_m + 4^i} \right\rfloor
\] actions. Hence, the total number of actions scheduled across all channels during the first $4^i K \tau$ rounds is
\begin{equation*}
\begin{aligned}
    \sum\limits_{m=1}^{M} \left\lfloor \frac{4^i K \tau}{\alpha_m + 4^i} \right\rfloor ~\geq~ \sum\limits_{m=1}^{M} \left( \frac{4^i K \tau}{\alpha_m + 4^i} - 1 \right) ~=~ K\tau \sum\limits_{m=1}^{M} \frac{1}{\alpha_m/4^i + 1} - M ~=~ K-M.
\end{aligned}
\end{equation*}
A lower bound of $(K-M)^+$ follows by the non-negativity of the number of scheduled pulls.\\
\textbf{Step B.} The second step is to show that the remaining number of actions $\hat{K} \leq K - (K-M)^{+} = \min{(K,M)}$ can be scheduled using an additional time of 
\[
    6 \left( \frac{\sum_{m=1}^{M} \alpha_m}{M} + 2\frac{4^i K}{M} \right).
\] Recall that Algorithm~\ref{scheduling-alg} divides the $4^i$ pulls of each of the remaining actions into \ifarxivFormat$\max{(1, \lfloor \frac{M}{2\hat{K}}}\rfloor)$ \else$\max{(1, \lfloor M / 2\hat{K}}\rfloor)$\fi equal parts and assign each part to an agent. Hence, each part will have number of pulls 
\begin{align}
    \frac{4^i}{\max{(1, \lfloor M / 2\hat{K}}\rfloor)} ~\overset{(a)}{\leq}~ \min{(4^i, \frac{4\hat{K}}{M} 4^i)} \label{eq:num_pulls_per_part}
\end{align}
and there will be at most $ M $ such parts. The first $\lfloor M / 2 \rfloor$ agents can be used for scheduling these parts in a way such that each agent is assigned at most three parts. It follows that each agent $m$ needs at most $3 \alpha_m + 3 \min{(4^i, \frac{4\hat{K}}{M} 4^i)}$ time to perform the scheduled pulls. Thus  the total number of rounds required to schedule the remaining pulls can be bounded by
\ifarxivFormat
\begin{align}
    3 \max_{m\in \{1,\cdots,\lfloor M/2 \rfloor\}}\alpha_m + 3 \min{(4^i, \frac{4\hat{K}}{M} 4^i)} &~\stackrel{(i)}{=}~ 3 \alpha_{\lfloor M/2 \rfloor} + 3 \min{(4^i, \frac{4\hat{K}}{M} 4^i)} \nonumber \\
    &~\stackrel{(ii)}{\leq}~ 6 \frac{\sum_{m=1}^{M} \alpha_m}{M} + 3 \min{(4^i, \frac{4\hat{K}}{M} 4^i)} \label{eq:endtime_second_part}
\end{align}
\else
\begin{align}
    3 \max_{m\in \{1,\cdots,\lfloor M/2 \rfloor\}}\alpha_m + 3 \min{(4^i, \frac{4\hat{K}}{M} 4^i)} &~\stackrel{(i)}{=}~ 3 \alpha_{\lfloor M/2 \rfloor} + 3 \min{(4^i, \frac{4\hat{K}}{M} 4^i)} ~\stackrel{(ii)}{\leq}~ 6 \frac{\sum_{m=1}^{M} \alpha_m}{M} + 3 \min{(4^i, \frac{4\hat{K}}{M} 4^i)} \label{eq:endtime_second_part}
\end{align}\fi
where $(i), (ii)$ follow from the fact that $\alpha_m$'s are ordered, i.e., $\alpha_1 \leq \alpha_2 \leq \cdots \leq \alpha_M$. Combining this with the result from \textbf{Step A}, we get that the end time needed to send all actions in batch $i$.

\subsection{Proof of Theorem 1}
\label{app:proof-thm-1}

\noindent{\textbf{\Theoref{main-thm}} \textit{
Consider a distributed multi-armed bandit setting  with $K$ actions and $M$ agents connected through heterogeneous erasure channels with erasure probabilities $\{ \epsilon_i\}_{i=1}^{M}$. If \algoname\ is run with horizon $T$, then the expected regret is,
    \begin{align*}
        \mathbb{E}[R_T] \leq c \Bigg( \Bigg. \sum_{a: \Delta_a > 0} \Big( \Big.  \frac{\log{(KMT)}}{\Delta_a} + \frac{M \log{(MT)}}{\sum\limits_{m=1}^{M} 1 / (\alpha_m + \frac{\log{(KMT)}}{\Delta_{a}})} \Big. \Big) + \sum_{m=1}^{M} \alpha_m \log{(MT)}  +  \log{(MT)} \Bigg. \Bigg)
    \end{align*}      
   where $\alpha_m = \lceil 4 \log{T} / \log{(1/\epsilon_m)}\rceil-1$ is the number of repetitions at agent $m$, $\Delta_a$ is the suboptimality gap for action $a$, and $c > 0$ a constant.}}

The regret bound is reached by bounding the number of batches a suboptimal arm can survive as a function of the suboptimality gap, conditioned on a good event that we specify later. This gives a bound on the maximum sub-optimality gap at each batch which in turn gives a bound on the regret using the bound on the batch length given in Lemma~\ref{lem:schedule-endtime}. 

Let $G$ be the event that for all agents $m$ and for all times $t$, at least one instruction among the times $t,t+1,\cdots,t+\alpha_m-1$ will not be erased. Hence, the event $G$ means that for any agent $m$, we cannot have $\alpha_m$ or more consecutive erasures. This implies that, conditioned on $G$, when an action $a$ is sent $\alpha_m + 4^i$ consecutive times by the learner to agent $m$, each of the last $4^i$ pulls will generate a reward from the distribution of action $a$. We call these last $4^i$ pulls, the effective pulls.  The probability of the compliment of $G$ can be bounded as
\begin{align}
\sP[G^c] \stackrel{(i)}{\leq} \sum_{m=1}^M \sum_{t=1}^T {\epsilon_m}^{\alpha_m} \stackrel{(ii)}{\leq} \sum_{m=1}^M \sum_{t=1}^T \frac{1}{T^4} \stackrel{(iii)}{\leq} \frac{1}{M T},
\end{align}
where $(i)$ follows by the union bound over all agents $m$ and times $t$, $(ii)$ uses \ifarxivFormat$\alpha_m = \lceil \frac{4 \log{T} }{ \log{(1/\epsilon_m)}} \rceil - 1$ \else$\alpha_m = \lceil 4 \log{T} / \log{(1/\epsilon_m)} \rceil - 1$\fi, and $(iii)$ follows from $M\leq T$.

Define an event $G'_i$ as
\begin{equation}
    G'_i = \left\{ | \evmu^{(j)}_a - \evmu_a| ~\leq~ 2 \sqrt{\frac{\log{(KMT)}}{2 \cdot 4^j}}  ~~\forall a \in \gA_j,~ j \in [i-1] \right\}, \nonumber
\end{equation}

where $\mu_a^{(j)}$ is the empirical mean calculated for action $a$ at batch $j$, as defined in step~\ref{step:emp-mean} in Algorithm~\ref{main}. By Hoeffding's inequality and the fact that rewards lie in $[0,1]$ almost surely, we have that $\sP [ G'_i | G] \geq 1 - 0.25 / {(MT)}$. Consequently, events $G$ and $G'_i$ happening together have a probability 
\begin{equation}\label{eq:GGi_high_prob}
    \sP [ G'_i \cap G] ~\geq~ (1-0.25 / (MT))^2 ~\geq~ 1 - 2 / (MT). 
\end{equation} 

We first bound the number of batches, a suboptimal arm can survive as a function of the suboptimality gap. Conditioned on $G \cap G'_{i+1} $  and the elimination criterion in Algorithm~\ref{main}, a sub-optimal action $a$ can survive getting eliminated in batch $i$ only if $4 \sqrt{\frac{\log{(KMT)}}{2 \cdot 4^i}} \geq \Delta_a / 2$.  This implies that $a$ can be in $\gA_{i+1}$ only when 
\begin{equation} \label{eq:bnd-elim-batch}
    i \leq \left\lceil \log_4{ \left( \frac{ 32 \log{(KMT)}}{\Delta_{a}^2}  \right)} \right\rceil,
\end{equation} i.e., whenever the batch number $i$ is greater than the bound provided in \eqref{eq:bnd-elim-batch}, $a \not\in \gA_i$. 

Using the result of Lemma~\ref{lem:schedule-endtime}, we know the number of sent instructions in each batch $i$ is upper bounded as
\begin{equation}\label{eq:n-pulls}
    M T^{(i)} ~\leq~ K^{(i)} M \cdot 4^i \tau + 6 \sum_{m=1}^{M} \alpha_m + 12 K^{(i)} 4^i,
\end{equation}
where $K^{(i)}=|\gA_i|$ is the number of actions at the start of batch $i$ and $T^{(i)}$ is the length of batch $i$. Conditioned on the event $G$ (we cannot have $\alpha_m$ consecutive erasures for any agent $m$), the last action played by agent $m$ in batch $i$ will be played at most $\alpha_m$ times in batch $i+1$ (due to potential erasures).
This implies that the total number of instructions, $T_i$, played by all agents due to instructions sent in batch $i$, can be bounded as 
\begin{equation}
T_i ~\leq~ \sum_{m=1}^{M} (T^{(i)} + \alpha_m) ~\leq~ K^{(i)}M \cdot 4^i \tau + 7  \sum_{m=1}^{M} \alpha_m + 12K^{(i)} 4^i .
\end{equation}

We utilize the following proposition, restated and proved at the end of \secref{app:proof-thm-1}, to bound the expected number of times a certain action is played due to an instruction sent in batch $i$.

\begin{restatable}{proposition}{propositionone}
 Conditioned on $(G\cap G'_i, \mathcal{A}_i)$, the expected number of times arm $a$ is played due to an instruction sent in batch $i$ is the same for all $a\in \mathcal{A}_i$. In particular, $\mathbb{E}[T_{ia}|G\cap G'_i,\mathcal{A}_i ]=\mathbb{E}[T_{ia'}|G\cap G'_i, \mathcal{A}_i], \quad  \forall a,a'\in \mathcal{A}_i$.
\end{restatable}

Conditioning on $\mathcal{A}_i$ in the previous proposition and in the following abbreviates conditioning on the event that the random set of surving actions in batch $i$ takes the value $\mathcal{A}_i$.


Then, we have that
\begin{align}\label{eq:n-pulls-per-arm}
    \mathbb{E}[T_{ia} | G\cap G'_i,\gA_i] ~=~ \frac{\mathbb{E}[T_{i} | G\cap G'_i, \gA_i]}{K^{(i)}} ~\leq~ M \cdot 4^i \tau + 7  \frac{\sum_{m=1}^{M} \alpha_m}{K^{(i)}} + 12 \cdot 4^i\ \ ~\forall a \in \mathcal{A}_i. 
\end{align}

Let $R_T^{(i)}$ be the regret of batch $i$. The regret of the algorithm can be bounded as
\ifarxivFormat
\begin{align}\label{eq:sep-reg}
    \mathbb{E}[R_T] &~=~ \sum\limits_{i=1}^{\log{(MT)}} \mathbb{E}[ R_T^{(i)}] ~\leq~ \sum\limits_{i=1}^{\log{(MT)}} \Big( \Big. \mathbb{E}[R_T^{(i)} |G \cap G'_{i}] + MT (1-\mathbb{P}[G \cap G'_{i}]) \Big. \Big) \nonumber \\
    &~\stackrel{(a)}{\leq}~ \sum\limits_{i=1}^{\log{(MT)}} (\mathbb{E}[\mathbb{E}[R_T^{(i)}|G \cap G'_{i}, \gA_i]] + 1) \nonumber \\
    &~=~ \sum\limits_{i=1}^{\log{(MT)}} \mathbb{E} [ \sum_a \mathbb{E}[T_{ia}|G \cap G'_{i}, \gA_i] \Delta_a ] + \log{(MT)} \nonumber \\
    &~\stackrel{(b)}{\leq}~ \sum\limits_{i=1}^{\log{(MT)}} \Bigg( \Bigg. \mathbb{E}[ \sum_a (N_1^{(i)}+N_2^{(i)}) \mathbb{E}[\mathds{1}[a\in \gA_i]|G \cap G'_{i}] \Delta_a] \nonumber \\
    & \qquad \qquad \qquad + \mathbb{E} [\sum_a 7 \sum_{m=1}^{M} \frac{\alpha_m}{K^{(i)}} \mathbb{E}[\mathds{1}[a\in \gA_i]|G \cap G'_{i}, \gA_i] \Delta_a ] \Bigg. \Bigg) + \log{(MT)} \nonumber \\
    &~\leq~ \sum\limits_{i=1}^{\log{(MT)}} \sum_a (N_1^{(i)}+N_2^{(i)}) \mathbb{E}[\mathds{1}[a\in \gA_i]|G \cap G'_{i}] \Delta_a + \mathbb{E} [7 \sum_{m=1}^{M} \alpha_m ] + \log{(MT)} \nonumber \\
    &~\leq~ \sum\limits_{i=1}^{\log{(MT)}} \sum_a (N_1^{(i)}+N_2^{(i)}) \mathbb{E}[\mathds{1}[a\in \gA_i]|G \cap G'_{i}] \Delta_a + c'' \log{(MT)} \sum_{m=1}^{M} \alpha_m + \log{(MT)},
\end{align}
\else
\begin{align}\label{eq:sep-reg}
    \mathbb{E}[R_T] &~=~ \sum\limits_{i=1}^{\log{(MT)}} \mathbb{E}[ R_T^{(i)}] ~\leq~ \sum\limits_{i=1}^{\log{(MT)}} \Big( \Big. \mathbb{E}[R_T^{(i)} |G \cap G'_{i}] + MT (1-\mathbb{P}[G \cap G'_{i}]) \Big. \Big) \nonumber \\
    &~\stackrel{(a)}{\leq}~ \sum\limits_{i=1}^{\log{(MT)}} (\mathbb{E}[\mathbb{E}[R_T^{(i)}|G \cap G'_{i}, \gA_i]] + 1) ~=~ \sum\limits_{i=1}^{\log{(MT)}} \mathbb{E} [ \sum_a \mathbb{E}[T_{ia}|G \cap G'_{i}, \gA_i] \Delta_a ] + \log{(MT)} \nonumber \\
    &~\stackrel{(b)}{\leq}~ \sum\limits_{i=1}^{\log{(MT)}} \Bigg( \Bigg. \mathbb{E}[ \sum_a (N_1^{(i)}+N_2^{(i)}) \mathbb{E}[\mathds{1}[a\in \gA_i]|G \cap G'_{i}] \Delta_a] \nonumber \\
    & & \mathllap{+ \mathbb{E} [\sum_a 7 \sum_{m=1}^{M} \frac{\alpha_m}{K^{(i)}} \mathbb{E}[\mathds{1}[a\in \gA_i]|G \cap G'_{i}, \gA_i] \Delta_a ] \Bigg. \Bigg) + \log{(MT)}} \nonumber \\
    &~\leq~ \sum\limits_{i=1}^{\log{(MT)}} \sum_a (N_1^{(i)}+N_2^{(i)}) \mathbb{E}[\mathds{1}[a\in \gA_i]|G \cap G'_{i}] \Delta_a + \mathbb{E} [7 \sum_{m=1}^{M} \alpha_m ] + \log{(MT)} \nonumber \\
    &~\leq~ \sum\limits_{i=1}^{\log{(MT)}} \sum_a (N_1^{(i)}+N_2^{(i)}) \mathbb{E}[\mathds{1}[a\in \gA_i]|G \cap G'_{i}] \Delta_a + c'' \log{(MT)} \sum_{m=1}^{M} \alpha_m + \log{(MT)},
\end{align}\fi
where $(a)$ follows from law of total expectation and \eqref{eq:GGi_high_prob}, $(b)$ follows from \eqref{eq:n-pulls-per-arm} and we use 
 $N_1^{(i)}= \frac{M}{\sum\limits_{m=1}^{M} 1 / (\alpha_m + 4^i)}$ and $N_2^{(i)}=12 \cdot 4^i$ for these quantities that do not depend on $\mathcal{A}_i$. We will bound each term in \eqref{eq:sep-reg} separately to get the final regret bound. 

We start by bounding the effect of the first term in~\eqref{eq:sep-reg}, $N_1^{(i)}=\frac{M}{\sum\limits_{m=1}^{M} 1 / (\alpha_m + 4^i)}$, on the final regret bound. We have that 
\begin{align}\label{eq:bnd-1}
    \sum\limits_{i=1}^{\log{(MT)}} \sum_a N_1^{(i)} \mathbb{E}[\mathds{1}[a\in \gA_i]|G \cap G'_{i}] \Delta_a &~=~ \sum\limits_{a} \sum\limits_{i=1}^{\log {(MT)}} \Delta_{a} \frac{M\mathbb{E}[\mathds{1}[a\in \gA_i]|G\cap G'_i]}{\sum\limits_{m=1}^{M} 1 / (\alpha_m + 4^i)} \nonumber \\
    &~\stackrel{(a)}{\leq}~ c \sum\limits_{a: \Delta_a > 0} \frac{M \log{(MT)}}{\sum\limits_{m=1}^{M} 1 / (\alpha_m + \frac{\log{(KMT)}}{\Delta_{a}})}
\end{align}
where $c$ is a universal constant, and $(a)$ follows from \eqref{eq:bnd-elim-batch} and the bound being an increasing function of $i$.

The effect of the second term in~\eqref{eq:sep-reg}, $N_2^{(i)}=12 \cdot 4^i$, 
\ifarxivFormat
\begin{align}\label{eq:bnd-3}
    \sum_a \sum\limits_{i=1}^{\log{(MT)}} N_2^{(i)} \mathbb{E}[\mathds{1}[a\in \gA_i]|G \cap G'_{i}] \Delta_a &~=~ 12 \sum\limits_{a} \sum\limits_{i=1}^{\left\lceil \log_4{ \left( \frac{ 32 \log{(KMT)}}{\Delta_{a}^2}  \right)} \right\rceil} 4^i \Delta_{a} \nonumber \\ &~\stackrel{(a)}{\leq}~ c' \sum_{a: \Delta_a > 0} \frac{\log{(KMT)}}{\Delta_a},
\end{align}
\else
\begin{align}\label{eq:bnd-3}
    & \sum_a \sum\limits_{i=1}^{\log{(MT)}} N_2^{(i)} \mathbb{E}[\mathds{1}[a\in \gA_i]|G \cap G'_{i}] \Delta_a ~=~ 12 \sum\limits_{a} \sum\limits_{i=1}^{\left\lceil \log_4{ \left( \frac{ 32 \log{(KMT)}}{\Delta_{a}^2}  \right)} \right\rceil} 4^i \Delta_{a} ~\stackrel{(a)}{\leq}~ c' \sum_{a: \Delta_a > 0} \frac{\log{(KMT)}}{\Delta_a}, 
\end{align}
\fi where $(a)$ follows from \eqref{eq:bnd-elim-batch}, and $c'$ is a universal constant. The final result follows by summing the bounds in \eqref{eq:bnd-1} and \eqref{eq:bnd-3}.

\propositionone*
\textit{Proof.} Recall that $T_{ia}$ is the number of times an agent plays arm $a$ due to an instruction sent in batch $i$.
We represent the schedule by the set $S = \{ \{ S_{mt} \}_{m=1}^{M} \}_{t=1}^{T^{(i)}}$, where $S_{mt}$ is the action the learner sends to agent $m$ at time $t$. Let $S(a\leftrightarrow a')$ represents the schedule where actions $a,a'$ are exchanged in the schedule $S$, i.e., $S(a\leftrightarrow a')_{mt}=a$ whenever $S_{mt}=a'$, $S(a\leftrightarrow a')_{mt}=a'$ whenever $S_{mt}=a$, otherwise $S(a\leftrightarrow a')_{mt}=S_{mt}$. We notice that conditioned on the schedule $S$ in batch $i$, whether an action is played in slot $t$ due to an instruction sent in batch $i$ is only a function of the erasures in batches $i,i+1,..$. Hence, we have that
\ifarxivFormat
\begin{align}
\mathbb{E}[T_{ia} | G \cap G_i', \mathcal{A}_i] &=~ \sum\limits_{S \in \mathbb{S}} \mathbb{P}[S | G \cap G_i', \mathcal{A}_i] \mathbb{E}[ T_{ia} | G \cap G_i',\mathcal{A}_i, S] =~ \sum\limits_{S \in \mathbb{S}} \mathbb{P}[S |\mathcal{A}_i] \mathbb{E}[ T_{ia} | G,S]\nonumber \\
&\stackrel{(a)}{=} \sum\limits_{S \in \mathbb{S}} \frac{1}{|\mathbb{S}|} \mathbb{E}[ T_{ia} |G, S] = \sum\limits_{S \in \mathbb{S}} \frac{1}{|\mathbb{S}|} \mathbb{E}[ T_{ia'} | G, S(a\leftrightarrow a')] = \sum\limits_{S \in \mathbb{S}} \frac{1}{|\mathbb{S}|} \mathbb{E}[ T_{ia'} | G, S] \nonumber \\ &=~ \mathbb{E}[T_{ia'} | G \cap G_i', \mathcal{A}_i],
\end{align}
\else
\begin{align}
\mathbb{E}[T_{ia} | G \cap G_i', \mathcal{A}_i] &=~ \sum\limits_{S \in \mathbb{S}} \mathbb{P}[S | G \cap G_i', \mathcal{A}_i] \mathbb{E}[ T_{ia} | G \cap G_i',\mathcal{A}_i, S] =~ \sum\limits_{S \in \mathbb{S}} \mathbb{P}[S |\mathcal{A}_i] \mathbb{E}[ T_{ia} | G,S]\nonumber \\
&\stackrel{(a)}{=} \sum\limits_{S \in \mathbb{S}} \frac{1}{|\mathbb{S}|} \mathbb{E}[ T_{ia} |G, S] = \sum\limits_{S \in \mathbb{S}} \frac{1}{|\mathbb{S}|} \mathbb{E}[ T_{ia'} | G, S(a\leftrightarrow a')] = \sum\limits_{S \in \mathbb{S}} \frac{1}{|\mathbb{S}|} \mathbb{E}[ T_{ia'} | G, S] =~ \mathbb{E}[T_{ia'} | G \cap G_i', \mathcal{A}_i],
\end{align}
\fi where $\mathbb{S}$ is the set of all (non-zero probability) possible schedules for batch $i$, and $(a)$ follows since the randomization in Algorithm~\ref{scheduling-alg} makes all the schedules in $\mathbb{S}$ equally probable.

\subsection{Proof of Theorem 2}
\label{app:proof-thm-2}

\noindent{\textbf{\Theoref{instance-indep-thm}} \textit{Consider the distributed multi-armed bandit setting with $K$ actions and $M$ agents connected through heterogeneous erasure channels $\{ \epsilon_i\}_{i=1}^{M}$. If \algoname is run for horizon $T$, then the expected regret is
    \begin{align*}
        R_T \leq c \Bigg( \Bigg. M \sqrt{\frac{ K T \log{(MT)} }{\sum_{m=1}^{M} 1 / (\alpha_m\Delta_{\star} + \log{(KMT)})}} ~ + \sum_{m=1}^{M} \alpha_m \log{(KMT)}\Bigg. \Bigg)
    \end{align*}      
   where $\Delta_{\star}$ is the value satisfying \[
    \Delta_{\star} = \frac{c' K \log{(MT)}}{T \sum\limits_{m=1}^{M} 1 / (\alpha_m + \frac{\log{(KMT)}}{\Delta_{\star}})},
   \] which can be efficiently approximated using the bisection method, $\alpha_m = \lceil 4 \log{T} / \log{(1/\epsilon_m)}\rceil -1$ number of repetitions and $c, c' > 0$ constants.}}

\noindent{\textit{Proof of \Theoref{instance-indep-thm}.}
From \eqref{eq:sep-reg}, the expected regret can be bounded as
\ifarxivFormat
\begin{align}\label{eq:ub_on_regret}
    \mathbb{E}[R_T]&~\leq~ \sum\limits_{i=1}^{\log{(MT)}} \sum_a \mathbb{E}[T_{ia}|G \cap G'_{i}] \Delta_a + \log{(MT)}\nonumber \\
    &~\leq~ MT \Delta + \sum\limits_{i=1}^{\log{(MT)}} \sum_{a:\Delta_a>\Delta} \mathbb{E}[T_{ia}|G \cap G'_{i}] \Delta_a + \log{(MT)} \nonumber \\
    &~\stackrel{(a)}{\leq}~ MT \Delta + \sum\limits_{i=1}^{\log{(MT)}} \sum_{a:\Delta_a>\Delta} (N_1^{(i)}+N_2^{(i)}) \mathds{1}[a\in \gA_i] \Delta_a + c''' \log{(MT)} \sum_{m=1}^{M} \alpha_m + \log{(MT)} \nonumber \\
    & ~\stackrel{(b)}{\leq}~ MT \Delta + c \sum\limits_{a: \Delta_a > \Delta} \left( \frac{M \log{(MT)}}{\sum\limits_{m=1}^{M} 1 / (\alpha_m + \frac{\log{(KMT)}}{\Delta_{a}})} + \frac{\log{(KMT)}}{\Delta_a} \right) + f(M, T, \valpha) \nonumber \\
    & ~\stackrel{(c)}{\leq}~ MT \Delta + c \sum\limits_{a: \Delta_a > \Delta}  \frac{2 M \log{(MT)}}{\sum\limits_{m=1}^{M} 1 / (\alpha_m + \frac{\log{(KMT)}}{\Delta_{a}})} + f(M, T, \valpha) \nonumber \\
    & ~\stackrel{(d)}{=}~ MT \Delta + \frac{c' K M \log{(MT)}}{\sum\limits_{m=1}^{M} 1 / (\alpha_m + \frac{\log{(KMT)}}{\Delta})} + f(M, T, \valpha) \nonumber \\
    &~\leq~ 2 \max{\left\{ TM \Delta, \frac{c' K M  \log{(MT)}}{\sum\limits_{m=1}^{M} 1 / (\alpha_m + \frac{\log{(KMT)}}{\Delta})} \right\}} + f(M, T, \valpha) ~~\forall \Delta > 0 
\end{align}
\else
\begin{align}\label{eq:ub_on_regret}
    \mathbb{E}[R_T]&~\leq~ \sum\limits_{i=1}^{\log{(MT)}} \sum_a \mathbb{E}[T_{ia}|G \cap G'_{i}] \Delta_a + \log{(MT)}\nonumber \\
    &~\leq~ MT \Delta + \sum\limits_{i=1}^{\log{(MT)}} \sum_{a:\Delta_a>\Delta} \mathbb{E}[T_{ia}|G \cap G'_{i}] \Delta_a + \log{(MT)} \nonumber \\
    &~\stackrel{(a)}{\leq}~ MT \Delta + \sum\limits_{i=1}^{\log{(MT)}} \sum_{a:\Delta_a>\Delta} (N_1^{(i)}+N_2^{(i)}) \mathds{1}[a\in \gA_i] \Delta_a + c''' \log{(MT)} \sum_{m=1}^{M} \alpha_m + \log{(MT)} \nonumber \\
    & ~\stackrel{(b)}{\leq}~ MT \Delta + c \sum\limits_{a: \Delta_a > \Delta} \left( \frac{M \log{(MT)}}{\sum\limits_{m=1}^{M} 1 / (\alpha_m + \frac{\log{(KMT)}}{\Delta_{a}})} + \frac{\log{(KMT)}}{\Delta_a} \right) + c''' \log{(MT)} \sum_{m=1}^{M} \alpha_m + \log{(MT)} \nonumber \\
    & ~\stackrel{(c)}{\leq}~ MT \Delta + c \sum\limits_{a: \Delta_a > \Delta}  \frac{2 M \log{(MT)}}{\sum\limits_{m=1}^{M} 1 / (\alpha_m + \frac{\log{(KMT)}}{\Delta_{a}})}  + c''' \log{(MT)} \sum_{m=1}^{M} \alpha_m + \log{(MT)} \nonumber \\
    & ~\stackrel{(d)}{=}~ MT \Delta + \frac{c' K M \log{(MT)}}{\sum\limits_{m=1}^{M} 1 / (\alpha_m + \frac{\log{(KMT)}}{\Delta})} + c''' \log{(MT)} \sum_{m=1}^{M} \alpha_m + \log{(MT)} \nonumber \\
    &~\leq~ 2 \max{\left\{ TM \Delta, \frac{c' K M  \log{(MT)}}{\sum\limits_{m=1}^{M} 1 / (\alpha_m + \frac{\log{(KMT)}}{\Delta})} \right\}} + c''' \log{(MT)} \sum_{m=1}^{M} \alpha_m + \log{(MT)} ~~\forall \Delta > 0 
\end{align}\fi where \ifarxivFormat $f(M, T, \valpha) = c''' \log{(MT)} \sum_{m=1}^{M} \alpha_m + \log{(MT)}$  and \else \fi$c, c', c''' > 0$ some constants. (a) follows from \eqref{eq:n-pulls-per-arm} where $N_1^{(i)}=M/(\sum_{m=1}^{M} 1 / (\alpha_m + 4^i))$ and $N_2^{(i)}=12 \cdot 4^i$. (b) follows from directly substituting \eqref{eq:bnd-1} and \eqref{eq:bnd-3} for the terms; and (c) follows from the fact that the first term is an increasing function of $\alpha_m$'s; therefore, 
$$\frac{M \log{(MT)}}{\sum\limits_{m=1}^{M} 1 / (\alpha_m + \frac{\log{(KMT)}}{\Delta_{a}})} ~\geq~ \frac{M \log{(MT)}}{\sum\limits_{m=1}^{M} 1 / \frac{\log{(KMT)}}{\Delta_{a}}} ~\geq~ \frac{\log{(KMT)}}{\Delta_a} ~~~\forall \{ \alpha_m\}_{m=1}^{M} \geq 0 . $$ (d) follows from the term inside the summation being a decreasing function of $\Delta_a$.

We choose $\Delta$ to be the value that minimizes the bound. Hence the optimal value $\Delta_{\star}$ satisfies:
\begin{align}
    TM \Delta_{\star} = \frac{c' K M \log{(MT)}}{\sum\limits_{m=1}^{M} 1 / (\alpha_m + \frac{\log{(KMT)}}{\Delta_{\star}})} \label{eq:optimal_delta}
\end{align}
Substituting \eqref{eq:optimal_delta} to the bound in \eqref{eq:ub_on_regret}, we get that
\begin{align}
    \mathbb{E}[R_T] & ~\leq~ 2 M \sqrt{ \frac{c' K T \log{(MT)}}{ \sum\limits_{m=1}^{M} 1 / (\alpha_m \Delta_{\star} + \log{(KMT)}) } } + c''' \log{(MT)} \sum_{m=1}^{M} \alpha_m + \log{(MT)}.
\end{align}

\bibliographystyle{IEEEtran}
\bibliography{references}

\end{document}